\pdfoutput=1

\documentclass[journal]{IEEEtran}
\usepackage{multirow}

\usepackage{cite}

\ifCLASSINFOpdf
  \usepackage[pdftex]{graphicx}
\else
\fi
\usepackage{amsmath}
\usepackage{array}

\usepackage{stfloats}
\usepackage{url}

\begin{document}
%
%
%
%

\title{Recognition of Visually Perceived Compositional Human Actions by Multiple Spatio-Temporal Scales Recurrent Neural Networks}
%
%
%


\author{Haanvid~Lee,
	Minju~Jung,
	and~Jun~Tani%


\thanks{H. Lee is with the Department
	of Electrical Engineering, Korea Institute of Science and Technology, Daejeon 305-701, Republic of Korea, e-mail: (haanvidlee@gmail.com).}
\thanks{M. Jung is with the Department
	of Electrical Engineering, Korea Institute of Science and Technology, Daejeon 305-701, Republic of Korea, e-mail: (minju5436@gmail.com).}
\thanks{J. Tani is with the Department
	of Electrical Engineering, Korea Institute of Science and Technology, Daejeon 305-701, Republic of Korea, and also with the Okinawa Institute of Science and Technology, Okinawa 904-0412, Japan, e-mail: (tani1216jp@gmail.com).}}

\maketitle

\begin{abstract}
The current paper proposes a novel neural network model for recognizing visually perceived human actions. The proposed multiple spatio-temporal scales recurrent neural network (MSTRNN) model is derived by introducing multiple timescale recurrent dynamics to the conventional convolutional neural network model. One of the essential characteristics of the MSTRNN is that its architecture imposes both spatial and temporal constraints simultaneously on the neural activity which vary in multiple scales among different layers. As suggested by the principle of the upward and downward causation, it is assumed that the network can develop meaningful structures such as functional hierarchy by taking advantage of such constraints during the course of learning. To evaluate the characteristics of the model, the current study uses three types of human action video dataset consisting of different types of primitive actions and different levels of compositionality on them. The performance of the MSTRNN in testing with these dataset is compared with the ones by other representative deep learning models used in the field. The analysis of the internal representation obtained through the learning with the dataset clarifies what sorts of functional hierarchy can be developed by extracting the essential compositionality underlying the dataset. 
\end{abstract}


%
\IEEEpeerreviewmaketitle

\section{Introduction}
%
%
%
%
\IEEEPARstart{R}{ecently,} a convolutional neural network (CNN) \cite{CNN}, inspired by a mammalian visual cortex, showed a remarkably better object image recognition performance than conventional vision recognition schemes which employ elaborately hand-coded visual features. A CNN trained with 1 million visual images from ImageNet \cite{CNN-ImageNet} was able to classify hundreds of object images with an error rate of 6.67\% \cite{CNN-6.67}, and demonstrated near-human performance \cite{CNN-HumanLevel}. However, CNNs lack the capacity for temporal information processing. As a consequence, CNNs are less effective in handling video image patterns than static images. 

To address this shortcoming, a number of action recognition models have been developed. Typical deep learning models for action recognition are 3D convolutional neural networks (3D-CNNs) \cite{3DCNN1}, long-term recurrent convolutional networks (LRCNs) \cite{LRCN}, and two-stream convolutional networks \cite{ActionRecog1-TwoStream}. The 3D-CNN extracts spatio-temporal features of videos through convolutions in the temporal and spatial domains in a fixed window \cite{3DCNN1}. The LRCN is a two-stage model that first extracts spatial features in its CNN stage and then extracts temporal features from its long-short term memory (LSTM) \cite{LSTM} stage \cite{LSTM}. And the two-stream convolutional network has one CNN stream for an RGB input, and the other CNN stream for an input of stacked optical flows. The two streams are joined at the end to make a categorical output \cite{ActionRecog1-TwoStream}.

Although the 3D-CNN, LRCN, and two-stream convolutional network perform well, some of their dynamics are not consistent with neuroscientific evidences. One important piece of evidence in mammals is that the size of the spatio-temporal receptive field of each cell is increased as the level goes higher \cite{s-t-field-mammal-increase1}, \cite{s-t-field-mammal-increase2}. And a principle from cybernetics era, so-called the downward causation \cite{Downward-Causation}, \cite{Downward-Causation-Gazzaniga} suggests that a spatio-temporal hierarchy can be naturally developed in human brains by taking advantage of macroscopic constraints genetically assigned to them. The evidence and principle suggest that a deep learning model for action recognition should form a hierarchy by assignment of spatio-temporal constraints. And the model should extract spatial and temporal features simultaneously in their hierarchy. But the representative action recognition models lack such properties.

The current study is an extension of the prior study using so-called the multiple spatio-temporal scales neural network (MSTNN) \cite{MSTNN}. The neural activity in the MSTNN is governed by both spatial and temporal constraints by means of local connectivity of convolutional layers and time constants assigned to the leaky integrator neural units at each layer \cite{MSTNN} respectively. These constraints make the MSTNN to develop faster and more local interaction in the lower layer whereas it develops slower and more global interactions in the higher level. This enables the MSTNN to extract spatio-temporal features in multiple levels from the exemplar data patterns. This formation of a spatio-temporal hierarchy is consistent with the biological evidence and the principle mentioned earlier.

However, it is true that the temporal processing capacity of the MSTNN is quite limited by the fact that its essential dynamics is a decay dynamics exerted by leaky integrator neurons \cite{MSTNN} that compose the model. At the same time, the MSTNN uses only forward connectivity without any recurrent connectivity which is considered to be the source of generating arbitrary complex dynamics, as known in the study of biological brains \cite{goel2014timing}. In this context, the current study attempts to add recursive dynamics to the MSTNN by introducing recurrent connectivity in the convolutional layers. This leads to our novel proposal of multiple spatio-temporal scales recurrent neural network (MSTRNN) model in the current study.

In the experiments, MSTRNNs were compared with MSTNNs and LRCNs. MSTRNNs were compared to the MSTNNs to observe how the existence of recurrent structures of MSTRNNs have effect on their action recognition performances. Among the introduced representative action recognition models, the LRCN was also used as a baseline of the MSTRNN since both have CNN and recurrent neural network structures. Also, it was chosen as a baseline model of the MSTRNN to see the effect of setting spatio-temporal constraints and extracting spatial and temporal features simultaneously. 

The MTSRNN model was evaluated by using three different human action datasets that are distinct in types and levels of compositionality introduced to the action patterns. In the first experiment, the MSTRNN is compared with the MSTNN on a dataset that was prepared by concatenating three action patterns that appear in Weizmann dataset \cite{Weizmann}. The experiment shows how the recurrent structure implemented on the MSTRNN is effective in action recognition task. Also a qualitative analysis on the neural activations generated from both models are made to show how their categorical memories are formed. The second and the third experiments were aimed to conduct comparative experiments using datasets of more natural human action patterns by controlling the level of the underlying compositionality. Because there were no public video datasets containing natural human actions with control of their compositionality level, two human action datasets were prepared for the current study. For the second experiment, a video dataset was built by considering compositions of objects and actions directed to those objects. In the experiment, the performance of the MSTRNN, MSTNN, and LRCN were compared. And analysis was made on the internal dynamics of the MSTRNN and LRCN by observing their time series activation patterns. The analysis was made to see how the multiple timescale constraint assigned to the MSTRNN can strengthen the model performance over the LRCN that does not involve with such temporal constraints. For the third experiment, triad compositions among a set of objects, object-directed actions, and modifiers for those actions are considered to prepare the dataset. The MSTRNN, MSTNN, and LRCN were again compared in the experiment. The experiment was designed to require the models to extract temporal features with longer temporal correlation for recognition of action modifiers than the one for objects and actions. By examining the recognition accuracies among these models, the advantage of the proposed model is clarified.

\section{PROPOSED MODEL}

The MSTRNN model consists of the following four types of layers: input, context, fully-connected, and an output layer as shown in Fig. 1 (A). The MSTRNN receives RGB image sequences in the input layer. Then from the image sequences, multiple context layers extract spatio-temporal features from the input image sequences. The extracted spatio-temporal features go through several layers of fully-connected layers. Then finally, in the output layer, categorical output is made. The output layer consists of softmax neurons.

The context layer of a MSTRNN simultaneously extracts spatio-temporal features, and is the core building block of the MSTRNN. The context layer consists of feature units, pooling units, and context units as shown in Fig. 1 (B). Each context layer is assigned a time constant that controls the decay dynamics of the context units and the feature units. A larger time constant causes the internal states of the leaky integrator neurons in the context layer to change more slowly at each time step \cite{MTRNN}. The MSTRNN assigns a larger time constant to higher layers to develop a spatio-temporal hierarchy \cite{MSTNN}, \cite{MTRNN}.

	\begin{figure}[!h] 
		\includegraphics[scale=0.55]{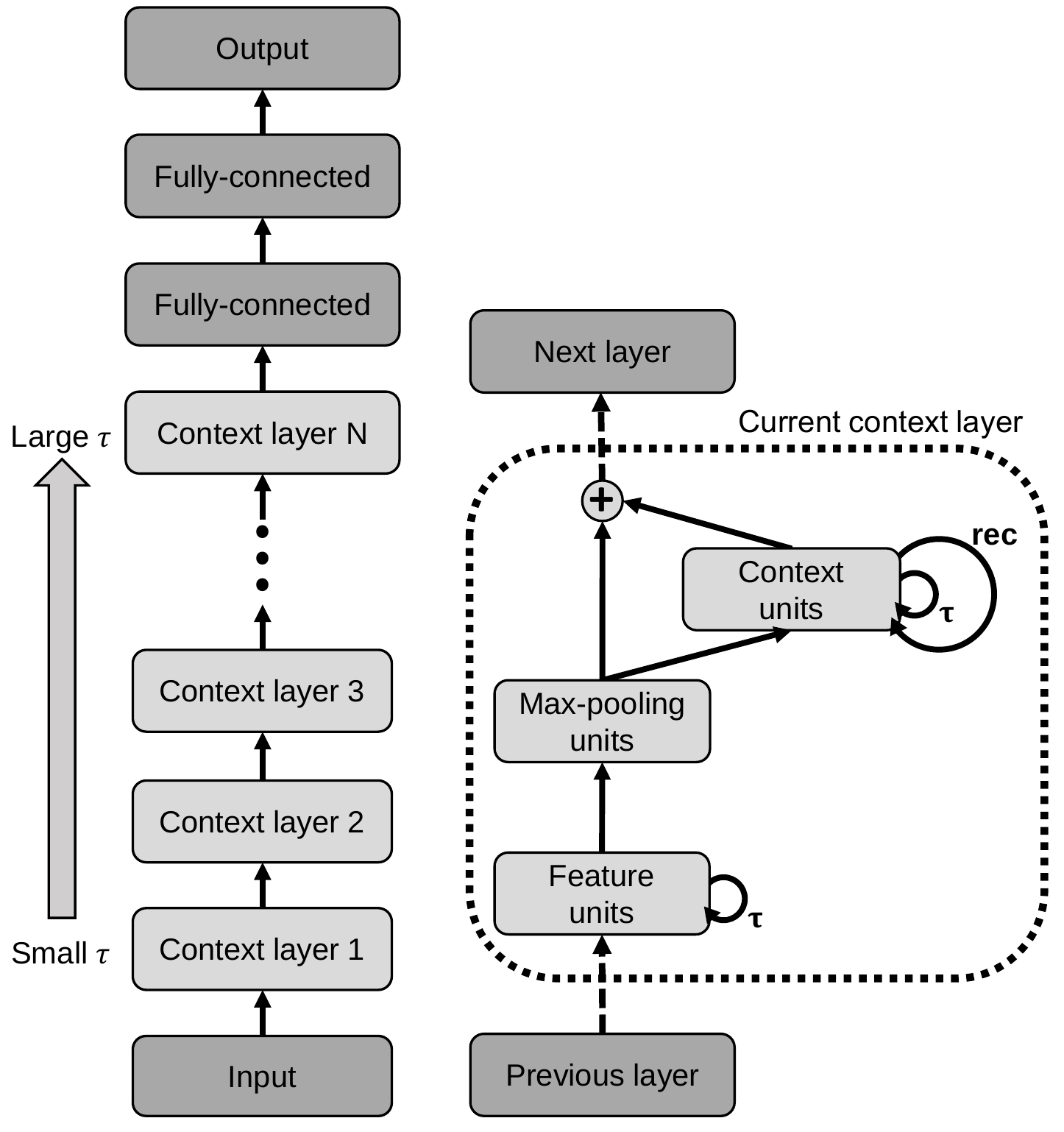}
		\centering
		\\ \qquad\qquad (A) \quad\qquad \qquad\qquad \qquad (B) \qquad\qquad\qquad
		\caption{{ The architecture of the MSTRNN.}
			(A) The full architecture of the MSTRNN. $\tau$ is a time constant for a set of leaky integrator neurons in a context layer. (B) The structure of the context layer. An arrow labeled $\tau$ indicates the decay dynamics of the leaky integrator neurons in the feature units and context units. The \textit{rec} with the arrow indicates recurrent connections made in the context units.}
		\label{fig1-MSTRNN_architecture}
	\end{figure}

\subsection{Feature Units}
A set of feature units (Fig. 1 (B)) is equivalent to a convolutional layer in CNN \cite{CNN}  composed of leaky integrator neurons with time constants to control their decay dynamics \cite{MSTNN}. The feature units are capable of extracting temporal features via the decay dynamics of the leaky integrator neurons, as well as being capable of extracting spatial features by convolutional operations. Feature units extract features from context units and pooling units of the previous context layer as shown in Fig. 1 (B), via convolutional kernels. The forward dynamics of the feature units are explained in (1) and (2). The internal state and the activation value of a neuron at the \textit{l}th context layer, \textit{m}th map of feature units, retinotopic coordinates \textit{(x, y)}, and   at time step \textit{t} are    represented as $\hat{f}^{txy}_{lm}$ and $f^{txy}_{lm}$  respectively.

\begin{align}
\hat{f}^{txy}_{lm} &= \Big(1-\frac{1}{\tau_l}\Big)\hat{f}^{(t-1)xy}_{lm}\\                                       &+ \frac{1}{\tau_l}\Big(\sum_{n=1}^{N_{l-1}}(k_{lmn} * p^{t}_{(l-1)n})^{xy}  + b_{lm}\Big)\nonumber\\                       &+ \frac{1}{\tau_l}\sum_{a=1}^{A_{l-1}}(z_{lma}*c^{t}_{(l-1)a})^{xy}\nonumber
\end{align}

\begin{eqnarray}
\label{eq:2}
f^{txy}_{lm} = 1.7159\tanh\Big(\frac{2}{3}\hat{f}^{txy}_{lm}\Big)
\end{eqnarray}

In (1) and (2), $\tau$ represents the time constant, \textit{k} and \textit{z} are the convolutional kernels that extract features from the previous layer pooling units and context units respectively. And \textit{b} is the bias used in the convolution operation, $*$ is the convolution operator, \textit{N} is the total number of maps of the pooling units, \textit{A} is the total number of maps of the context units. Additionally, \textit{p} and \textit{c} are the activation values of the pooling units and the context units respectively. The first term on the right hand side of (1) describes the decay dynamics of the leaky integrator neurons. The second term represents the convolution of the features in the pooling units. And, the third term describes the features extracted from the context units of the previous context layer. Equation (2) shows the hyperbolic tangent function \cite{LecunTanh} that is used as the activation functions of the feature units.

\subsection{Context Units}
A set of context units (Fig. 1 (B)) is equivalent to feature units with recurrent convolutional kernels. Due to the addition of recurrent convolutional kernels in the structure of the features units, it has more enhanced temporal processing capacity than the feature units. The recurrent dynamics of the context units enhance the extraction of latent temporal features from input image sequences \cite{Elman-RNN}, \cite{Jordan-RNN}. The recurrent connections are made by 3x3 convolutional kernels with 1 zero padding. The context units get inputs from themselves using the recurrent kernels and pooling units in the same layer (see Fig. 1 (B)) via convolutional kernels. The forward dynamics of the context units are shown in (3) and (4). The internal state and activation value of a neuron in \textit{(x, y)} coordinates of the \textit{a}th map of the context units in the \textit{l}th context layer, at time step \textit{t} are represented as  $\hat{c}^{txy}_{la}$ and $c^{txy}_{la}$ respectively.

	\begin{align}
	\label{eq:3}
	\hat{c}^{txy}_{la} &= \Big(1-\frac{1}{\tau_l}\Big)\hat{c}^{(t-1)xy}_{la}\\     &+ \frac{1}{\tau_l}\Big(\sum_{m=1}^{M_{l}}(\tilde{k}_{lam} * p^{t}_{lm})^{xy} + \tilde{b}_{la}\Big)\nonumber\\   &+ \frac{1}{\tau_l}\Big(\sum_{b=1}^{B_{l}}(\tilde{z}_{lab}*c^{t-1}_{lb})^{xy} \Big)\nonumber
	\end{align}

	\begin{eqnarray}
	\label{eq:4}
	c^{txy}_{la} = 1.7159\tanh\Big(\frac{2}{3}\hat{c}^{txy}_{la}\Big)
	\end{eqnarray}

	In (3) and (4), $\tau$ represents the time constant, $\tilde{k}$ is the convolutional kernel, $\tilde{b}$ is the bias for the convolution operation, $*$  is the convolution operator, \textit{M} is the total number of maps of the pooling units,  \textit{B} is the total number of maps of the context units,  $\tilde{z}$  is the recurrent convolutional kernel of the context units, \textit{p} and \textit{c} are the neural activations of pooling units and context units respectively. The first term on the right hand side of (3) describes the decay dynamics of the leaky integrator neurons. The second term represents the convolution of the pooling units. The third term describes the recurrent dynamics in terms of recurrent shared weights. The neural activations of the context units in the previous time step are supplied through the recurrent convolutional kernels. For the activation function of the context units, the MSTRNN uses the same hyperbolic tangent function that was used for the feature units as shown in (4) \cite{LecunTanh}.

	\subsection{Training Method}

	Training was conducted in a supervised manner using the delay response scheme \cite{MSTNN}. Black frames were input to the MSTRNN after each input image sequence during the delay response period. In this period, errors were calculated for each time step by comparing the outputs of the MSTRNN with the true labels of the input image sequences using the Kullback-Leibler divergence. The cost function used in the training phase is shown in (5). The error calculated for a whole input image sequence is represented as \textit{E}.

		\begin{eqnarray}
		\label{eq:5}
		E = \sum_{t=T+1}^{T+d}\sum_{s=1}^{S}\sum_{n=1}^{N_{s}} \tilde{o}_{sn} \ln \Big( \frac{\tilde{o}_{sn}}{o^t_{sn}} \Big)
		\end{eqnarray}

	In (5), \textit{d} is the delay response period, \textit{T} is the duration of an input video (length of frames), \textit{S} is the total number of softmax vectors in the output layer. The output layer can have several softmax vectors depending on a task. $N_{s}$ is the total number of neurons in the \textit{s}th softmax vector of the output layer, $\tilde{o}$  is the true output, and  \textit{o}  is the output categorized by the MSTRNN. The true output was given as the one-hot-vector for each softmax vector. Here, the term one-hot-vector refers to a vector of values where a value is $"1"$ for the one and only correct category and $"0"$ for the rest of the categories. 
	
	The error for each input action video is obtained with (5). The error is used to optimize the learnable parameters using the back propagation through time (BPTT) \cite{BPTT}, and the stochastic gradient descent algorithms. To prevent overfitting, all learnable parameters (except biases) were learned with a weight decay of 0.0005 \cite{WeightDecay}. In addition, 50\% dropout \cite{Dropout} and random cropping \cite{3DCNN2} of the input images that are 10 pixels smaller by width and height were also used to avoid overfitting. 
	
	Learning rates started from 0.01 and decayed by 6\% at every epoch for the first experiment. In the second and third experiments, learning rates decayed by 2\%. The other models that were used for comparison with the MSTRNN were also trained with this method.

	\subsection{Performance Evaluation}
	
	To evaluate categorization performance, the leave-one- subject-out cross-validation (LOSOCV) scheme was used. In this method, one subject was selected from the dataset and his/her video clips were left out of the training data, and were instead used as a test data. Accuracies obtained from all possible validation sets were averaged to be used as an evaluation measure. When there was more than one set of categories, such as actions and action-related objects, recognition accuracy for action-ADO pair was computed. Then the epoch where a model showed best accuracy on the joint category was picked. Then the accuracy of each set of category (objects or actions) at the epoch was used for evaluation. The accuracies were rounded off to the second decimal place, and recorded as measures of overall performance.
	
	\section{EXPERIMENTS}
	In the first experiment, the MSTRNN was compared with the MSTNN using a relatively simple action dataset that has a temporal compositionality. Then in the second and third experiments, the MSTRNN was tested on datasets that look more natural and have higher compositionality levels. In the second and third experiments, the MSTRNN was compared with the MSTNN and LRCN.
	
	\subsection{Model Parameter Settings}
	The MSTRNN was compared with MSTNN and MSTRNN in the experiments. All three models have same input and output layer according to the datasets they were tested. Table I shows the RGB input image sizes according to the datasets after cropping which is mentioned in the \textit{Training Method} section. Output layer configurations according to the datasets are also described in Table I. The three actions concatenated patterns from Weizmann Dataset (3ACWD), the compositionality level 1 action dataset (CL1AD), and the compositionality level 2 action dataset (CL2AD) were used in the first, second, and third experiments respectively. The parameter settings that differ by models are described below.
	
		\begin{table}[!ht]
			\centering
			\caption{
				{Input and Output Of The Tested Models}}
		      \begin{tabular}{cccr}
		      	\hline
		      	Dataset                & \begin{tabular}[c]{@{}c@{}}Input\\ size\end{tabular} & \begin{tabular}[c]{@{}c@{}}Output\\ category\end{tabular} & \multicolumn{1}{c}{\begin{tabular}[c]{@{}c@{}}Softmax\\ neurons\end{tabular}} \\ \hline
		      	3ACWD                  & 38x44                                                & Action                                                    & 27                                                                            \\ \hline
		      	\multirow{2}{*}{CL1AD} & \multirow{2}{*}{108x108}                             & Object                                                    & 4                                                                             \\
		      	&                                                      & Action                                                    & 9                                                                             \\ \hline
		      	\multirow{3}{*}{CL2AD} & \multirow{3}{*}{108x108}                             & Object                                                    & 4                                                                             \\
		      	&                                                      & Action                                                    & 4                                                                             \\
		      	&                                                      & Modifier                                                  & 6                                                                             \\ \hline
		      \end{tabular}
			\label{table1}
		\end{table}
		
		\textbf{MSTRNN}: The MSTRNN model has one convolutional layer, one pooling layer, two context layers, two fully-connected layers, and a softmax layer as described in Fig. 2 (A). The first and second context layers have time constants of 2 and 100 respectively. The convolutional layer and pooling layer were used before the context layers to decrease the size of input for the context layers to decrease the computing time. This is because a context layer takes more computing time than a convolutional layer and a pooling layer. With few exceptions, all convolutional and pooling kernels used in the model are 3x3 (3 pixels wide, 3 pixels high) with stride of 1 and 2x2 with stride of 2 respectively. These were used since they were found to be best for CNNs \cite{small_kernel_best_simonyan2014very}.
		
		When the model was tested on the 3ACWD, 3x5 kernels were used instead of 3x3 on the first convolutional layer and the set of convolutional kernels that were first used after the first pooling layer (see Fig. 2 (A)). In the experiment with compositionality level 1 action dataset (CL1AD) and compositionality level 2 action dataset (CL2AD), the first convolutional layer used 3x3 kernels with stride of 3. 
		
		Also, the exceptions were made on the convolutional kernels that are connected to context units (see Fig. 1 (B)). The recurrent convolutional kernels of context units have size of 3x3 with stride of 1 and 1 zero padding. The convolutional kernels that connect pooling units to context units and context units to the next layer have size of 2x2 with stride of 1. These exceptions were made so that the resulting map size of the context units are unchanged as a time step progresses.
		
		\textbf{Baselines}: Two baselines were used in our experiments. As one of the two baselines, the MSTNN were used for all three experiments.  The MSTNN model has one convolutional layer, two convolutional layers consisted of leaky integrator neurons, three pooling layers, two fully-connected layers, and a softmax layer (Fig. 2 (B)). The first and second convolutional layers consisted of leaky integrator neurons have time constants of 2 and 100 respectively. Also, the LRCN was used as a baseline in the experiments with CL1AD and CL2AD. It has three convolutional layers, three pooling layers, one LSTM layer, one fully-connected layer, and a softmax layer (Fig. 2 (C)). One LSTM layer was used following the works of Donahue et al. \cite{LRCN}. As in the MSTRNN parameter settings, with the same exceptions, all of the convolutional and pooling kernels of MSTNN and LRCN are 3x3 with stride of 1 and 2x2 with stride of 2 respectively due to the same reason as in the case of MSTRNN parameter settings. The MSTNN and LRCN models have the same exceptions of not using convolutional kernel of 3x3 with stride of 1 on their first convolutional layer and first set of convolutional kernels that are used after the first pooling layer. The parameters of MSTNNs and LRCNs were adjusted so that they have similar number of parameters to MSTRNNs. Table II shows the number of parameters used in the MSTRNN, MSTNN, and LRCN in the second experiment. The number of parameters in the three models differ between the experiments  that use different datasets. But the changes in the parameters of the models according to datasets mentioned above are very subtle and the number of parameters are kept similar to each other for all experiments.

			\begin{figure}[!ht]
				%
				\includegraphics[scale=0.55]{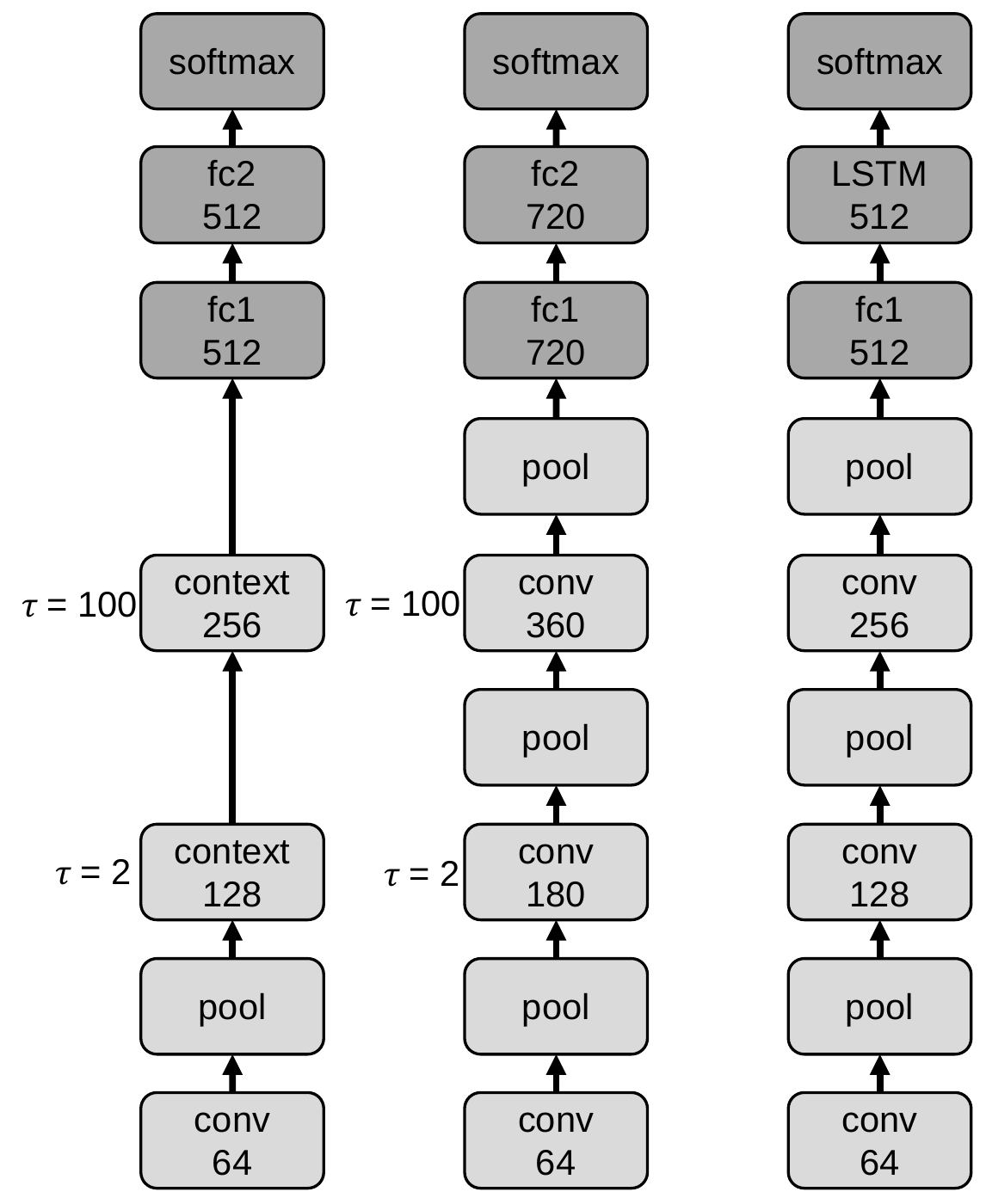}
				\centering
				\\\qquad \qquad\qquad  (A) \qquad \qquad (B) \qquad\qquad(C)\qquad\qquad\quad
				\caption{{ Architectures of the models used in the experiments. }
					(A) Architecture of the MSTRNN. (B) Architecture of the MSTNN. (C) Architecture of the LRCN.}
				\label{fig2}
				
			\end{figure}
			
					\begin{table}[!ht]
						\centering
						\caption{
							{ Number of parameters in the tested models}}
	\begin{tabular}{ccc}
		\hline
		MSTRNN & MSTNN & LRCN \\ \hline
		\multicolumn{1}{r}{3,540,365} & \multicolumn{1}{r}{3,549,433} & \multicolumn{1}{r}{3,655,053}
		\\\hline
	\end{tabular}
	
						\label{table1}
					\end{table}
					
\subsection{Categorization of Three Actions Concatenated Patterns from the Weizmann Dataset}

This section compares action categorization capability of the multiple spatio-temporal scales recurrent neural network (MSTRNN) and that of the multiple spatio-temporal scales neural network (MSTNN) \cite{MSTNN} with analyzing the internal dynamics developed in each model. For this purpose, three actions concatenated patterns were prepared by using Weizmann dataset \cite{Weizmann}. 

\textbf{Dataset}: A set of compositional action videos was prepared by concatenating videos of three different human actions from the Weizmann dataset \cite{Weizmann}. The three actions were jump-in-place (JP), one-hand-wave (OH), and two-hand-wave (TH) as shown in Fig. 3, resulting in 27 categories, and one video clip of the concatenated actions for each category. 27 videos for each of the nine subjects exist in the dataset. The foreground silhouettes of the resulting 3ACWD were emphasized by background subtraction utilizing background sequences, and were resized to 48x54.

			\begin{figure}[!h]
				
				\includegraphics[scale=0.3]{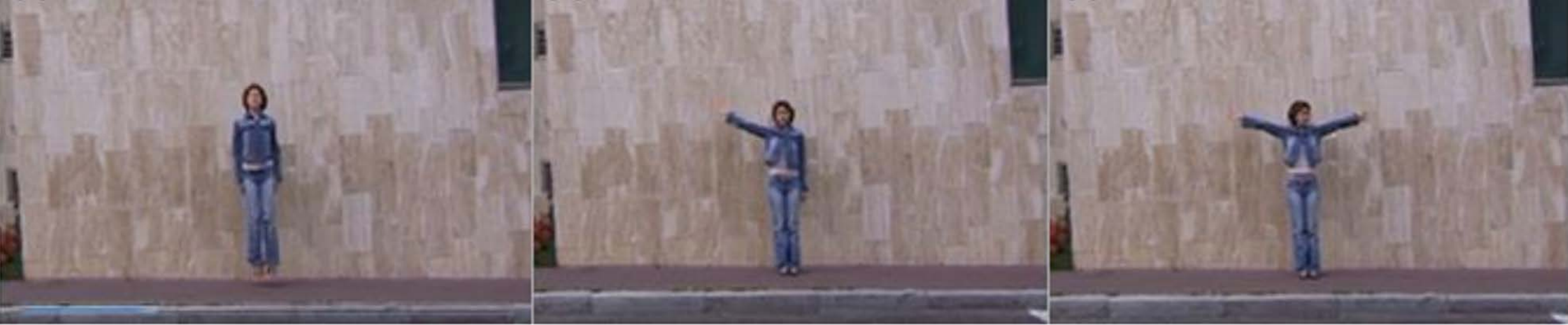}
				\centering
				\\ \quad  (A) \qquad \qquad\qquad(B) \qquad\qquad\qquad(C)\quad\qquad
				\caption{{ The 3 human action categories used from the Weizmann dataset.}
					(A) Jump-in-place. (B) One-hand-wave. (C) Two-hand-wave.}
				\label{fig3}
			\end{figure}
								
\textbf{Results}: The categorization accuracy of the MSTRNN on the 3ACWD was more than double the categorization accuracy obtained from MSTNN (Table III). This result implies that context units with recurrent weights improve the categorization of long concatenated human action sequences.

					\begin{table}[!ht]
						\centering
						\caption{
							{ Action recognition accuracies on 3ACWD in percentage}}
\begin{tabular}{@{}ccc@{}}
	\hline
	Category & MSTRNN & MSTNN \\ \hline
	Action & \multicolumn{1}{r}{92.3} & \multicolumn{1}{r}{45.27} \\ \hline
\end{tabular}
						
						\label{table1}
					\end{table}

		Next, the internal dynamics of MSTRNN and MSTNN were assessed by time series analysis of their neural activation values. The activation values were obtained from the third convolutional layer of the MSTNN  and the context units in the second context layer of the MSTRNN (see Fig. 2). The time series neural activations were visualized by using principle component analysis (PCA) \cite{PCA}. Only the first and the second principle components of the neural activations were used for the visualization. In the following discussion of this analysis, X indicates arbitrary primitive actions, and the time series of neural activations obtained when given action primitives A, B, C in a sequential manner is designated “PCA trajectory A-B-C”. Since both models are trained to categorize actions based on the outputs obtained during the teaching length, the end position of trajectories should be differentiated based on the history of images that were shown in a sequence.
		
		The PCA trajectories of the MSTNN (Fig. 4 (A)) at any given time step are largely affected by the primitive action of the input video at that time step. Accordingly, neural activations approached points representative of the current action in the PCA mapped space. Trajectories JP-X-X, OH-X-X, TH-X-X first approached points marked “JP”, “OH”, “TH” in Fig. 4 (A) respectively before the presentation of second and third primitives. When the second and third primitives were given, the trajectories immediately changed their directions toward the representative positions of the primitives that were recently shown. Therefore two branching points can be seen at each of the JP-X-X, OH-X-X, and TH-X-X trajectories.
		
		Unlike the characteristics shown for the MSTNN, the PCA trajectories obtained from the context units of the second context layer in the MSTRNN (Fig. 4 (B)) did not simply approach positions representative of currently displayed action primitives in the PCA mapped space, but tended to differentiate the primitives from each other.

\begin{figure*}[!ht]
	
	\includegraphics[scale=0.52]{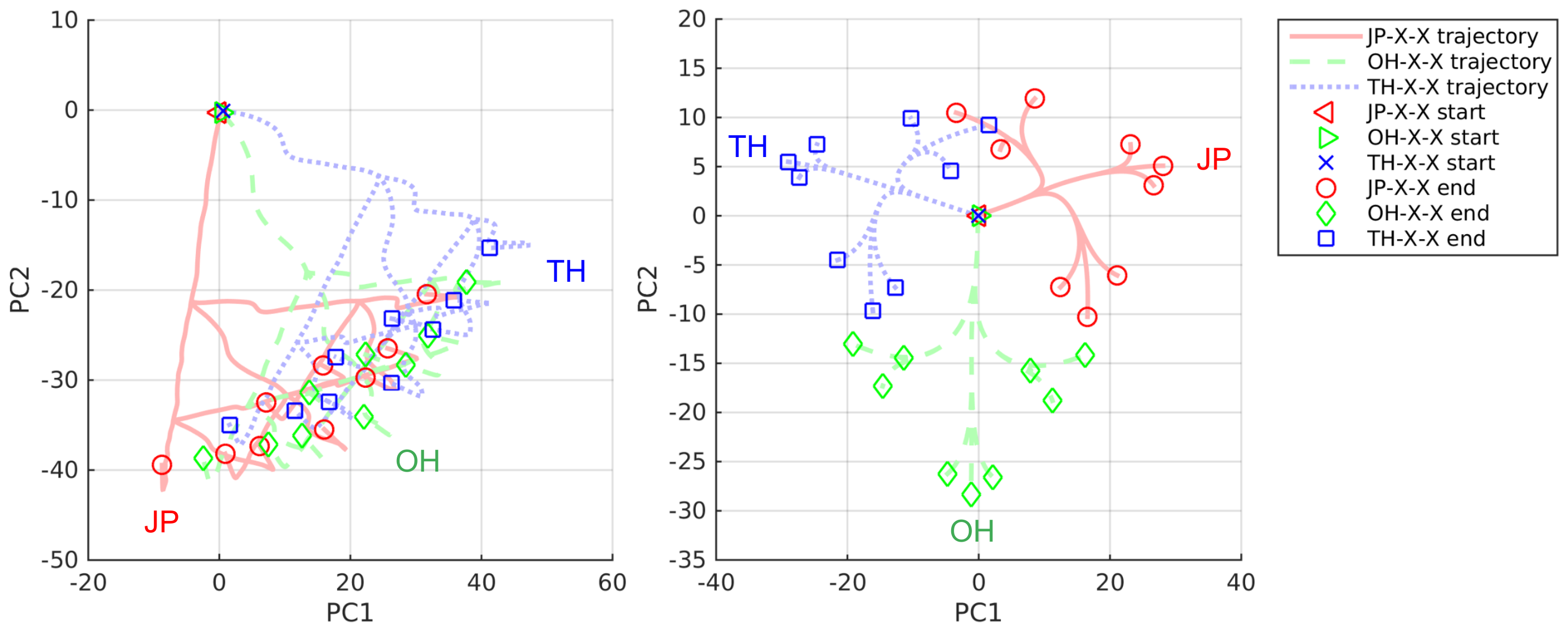}
	\centering
	\\ \qquad  (A) \qquad \qquad\qquad\qquad\qquad\qquad\qquad\qquad(B) \qquad\qquad\qquad\qquad\qquad
	\caption{{ PCA mapping of the time series activation values obtained from the MSTRNN and MSTNN when test data was given to the models.}
		(A) PCA mapping generated from the third convolutional layer of the MSTNN. (B) PCA mapping generated from the context units in the second context layer of the MSTRNN.}
	\label{fig3}
\end{figure*}				
\begin{figure*}[!ht]
	
	\includegraphics[scale=0.52]{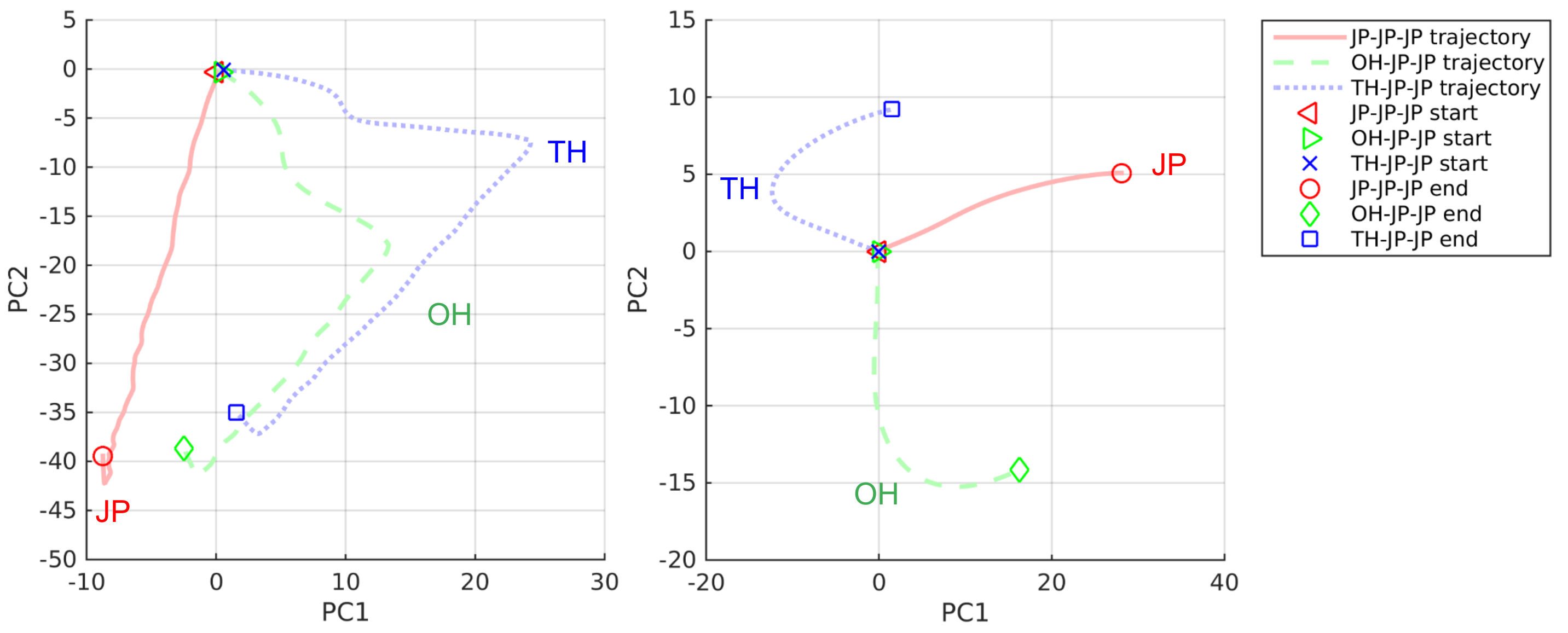}
	\centering
\\ \qquad  (A) \qquad \qquad\qquad\qquad\qquad\qquad\qquad\qquad(B) \qquad\qquad\qquad\qquad\qquad
	\caption{{ PCA mapping of the time series activation values obtained when the test subject$'$s action videos of the JP-JP-JP, OH-JP-JP, and TH-JP-JP concatenated action sequences were given to both the MSTNN and MSTRNN models.}
		(A) PCA mapping generated from the third convolutional layer of the MSTNN. (B) PCA mapping generated from the context units in the second context layer of the MSTRNN.}
	\label{fig3}
\end{figure*}								
		
To show the characteristics of the MSTNN and MSTRNN in more detail, PCA trajectories of JP-JP-JP, OH-JP-JP, and TH-JP-JP of the MSTNN and MSTRNN models are shown in Fig. 5. In the trajectories obtained from the MSTNN (Fig. 5 (A)), the JP-JP-JP trajectory directly approached the marker “JP” in the figure. And the trajectories OH-JP-JP and TH-JP-JP first approached regions marked “OH” and “TH” respectively. After the second and the third primitives (JP and JP), the trajectories changed their paths and approached the region marked “JP”. 

But the PCA trajectories obtained from the MSTRNN did not converge to the ”JP” position marked in the figure (Fig. 5 (B)). The JP-JP-JP trajectory of the MSTRNN directly approached the marker “JP” in the figure, same as the MSTNN. But for the trajectories of OH-JP-JP and TH-JP-JP, they first approached regions marked “OH” and “TH” respectively. Then, when the second (JP) and third primitives (JP) were shown, the trajectories changed their paths. However, the direction of the trajectories did not point toward the ”JP” position in the figure. 

This shows that the MSTRNN has better categorical memory than the MSTNN since it was able to keep the temporal information of the previously shown ”OH” and ”TH” action primitives, and prevent the activation values from becoming similar to the values obtained when the ”JP” action primitive was shown to the model. 

The decay dynamics of the MSTNN are responsible for the trajectories in Fig. 5 (A) approaching the markers which represent the current action primitives. As described in (1), the internal neural values of the feature units are affected by both the current spatial-temporal features processed in the previous context layer, and the decayed internal neural values of the units of the previous time step. In this method, previously extracted spatial features do not effectively influence the internal neural values of the current time step to keep track of which action primitives came in the past, since they gradually decay over time. 

The higher performance of categorical memories in the MSTRNN, compared to that of the MSTNN, is due to the recurrent structure in the context units. The context units retain important spatio-temporal features extracted during previous time steps, and are then more or less reinforced during current steps. And it can be inferred that the improved categorical memories of the MSTRNN led to better categorization performance compared to the MSTNN.

\subsection{Categorization of Compositionality Level 1 Action Dataset}
This experiment compared the categorization performance of the MSTRNN, MSTNN, and LRCN with the newly prepared compositionality level 1 action dataset (CL1AD). Each action pattern in CL1AD was generated by composition of a specific action and corresponding action-directed-object (ADO) in its category. In this experiment, the PCA representation of the activation values obtained from MSTRNN was observed and compared to the neural activations generated from LRCN. 

\textbf{Dataset}: CL1AD consists of 900 videos that are made by 10 subjects performing nine actions directed at four objects. There are 15 object-directed action categories in total (Fig. 6). The dataset was designed so that the categorization task was non-trivial. A non-ADO (distractor) appears along with an ADO in each video to prevent the model from inferring a human action or an ADO solely by recognizing an object in a video. For each object-directed action category, six videos were shot for each subject with three different non-ADOs appearing in two videos each in different states (opened or closed) if possible, as they are presented in the other videos as ADOs. During the recording of the dataset, the subjects generated each action without constraints. Objects were located in random positions in the task space. However, the camera view angle was fixed, since the problem of view invariance is beyond the scope of the current study. The dataset is open for public use (available at \url{http://neurorobot.kaist.ac.kr/programs.html}).

			\begin{figure}[!h]
				
				\includegraphics[scale=0.7]{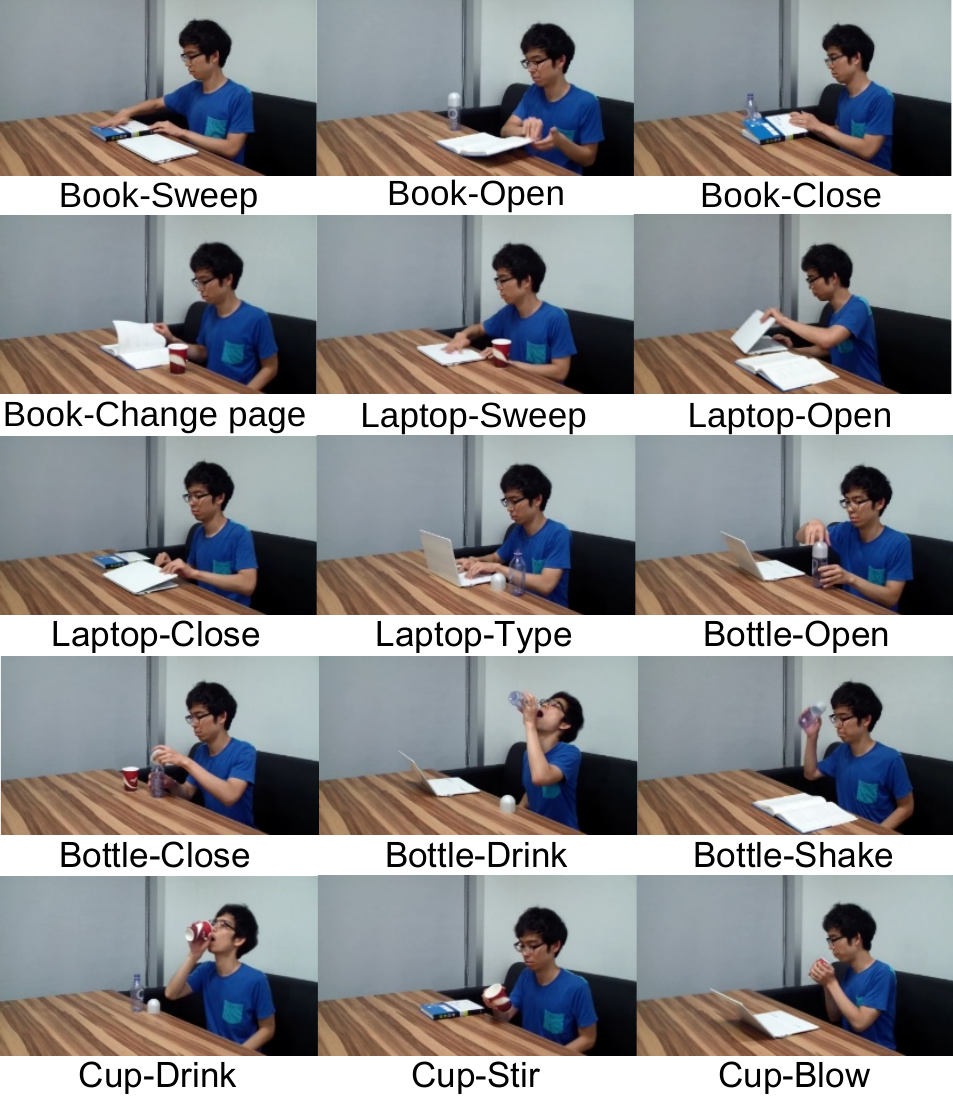}
				\centering
				\caption{{ Sample frames from CL1AD. An object that is not related to the action also appears in the scene to make the recognition task non-trivial.}}
				\label{fig6}
			\end{figure}
			
\textbf{Results}: The results obtained by testing both the MSTRNN and MSTNN on the CL1AD are shown in Table IV. The MSTRNN showed the highest action categorization performance among the models. This confirms that the MSTRNN, which are characterized by both of the multiscale spatio-temporal constraint and the recurrent connectivity, outperformed both of the MSTNN without the recurrent connectivity, and the LRCN without the multiscale temporal constraint. Although the LSTM, which is considered to have a capability of self-adapting timescale of own dynamics by adjusting the ratio of the forgetting gate, can perform well in generating and recognizing low dimensional or discrete sequence data \cite{chinese_hand_writing_lstm}, \cite{HandWritingBengio}, \cite{graves2013generating}, its performance cannot be guaranteed with extremely high dimensional temporal data such as video data without introducing simultaneous constraint for spatial processing and temporal one at each layer. Also, in the current result, MSTRNN and MSTNN showed similar ADO categorization accuracies that are higher than the one by LRCN. This might have resulted because both of the MSTRNN and MSTNN extracted spatio-temporal features simultaneously while the LRCN extracted spatial features and extract temporal features in different stages.

					\begin{table}[h]
						\centering
						\caption{
							{Categorization accuracies on CL1AD in percentage}}
\begin{tabular}{crrr}
	\hline
	Category & \multicolumn{1}{c}{MSTRNN} & \multicolumn{1}{c}{MSTNN} & \multicolumn{1}{c}{LRCN} \\ \hline
	ADO      & 86.89                      & 84                        & 66                       \\ \hline
	Action   & 77.44                      & 59.67                     & 58.44 \\ \hline                   
\end{tabular}
						
						\label{table1}
					\end{table}
					
Next, we observed how the temporal constraints in the MSTRNN helped with action categorization by comparing the time series activations of the MSTRNN and LRCN. Neural activations in the first fully-connected layer of the MSTRNN, and the LSTM layer of the LRCN (see Fig. 2), which are generated from the input action videos belonging to Close/Drink-Bottle, were visualized by PCA in a manner similar to the previous experiment.  

For the PCA trajectories obtained from the MSTRNN (Fig. 7 (A)), the trajectories share paths until some point, then branch out according to their action categories. This is because the bottle in Close/Drink-Bottle category first appears in their “opened” states, so that the image and movement in the videos are similar until a subject hand reaches out for the cup. The image and movement then start to differ after the objects are reached by the subject, and the subject starts to act according to an action category. The PCA trajectories seem to represent perceptual time series inputs in categorized space. 

\begin{figure*}[!ht]
	
	\includegraphics[scale=0.52]{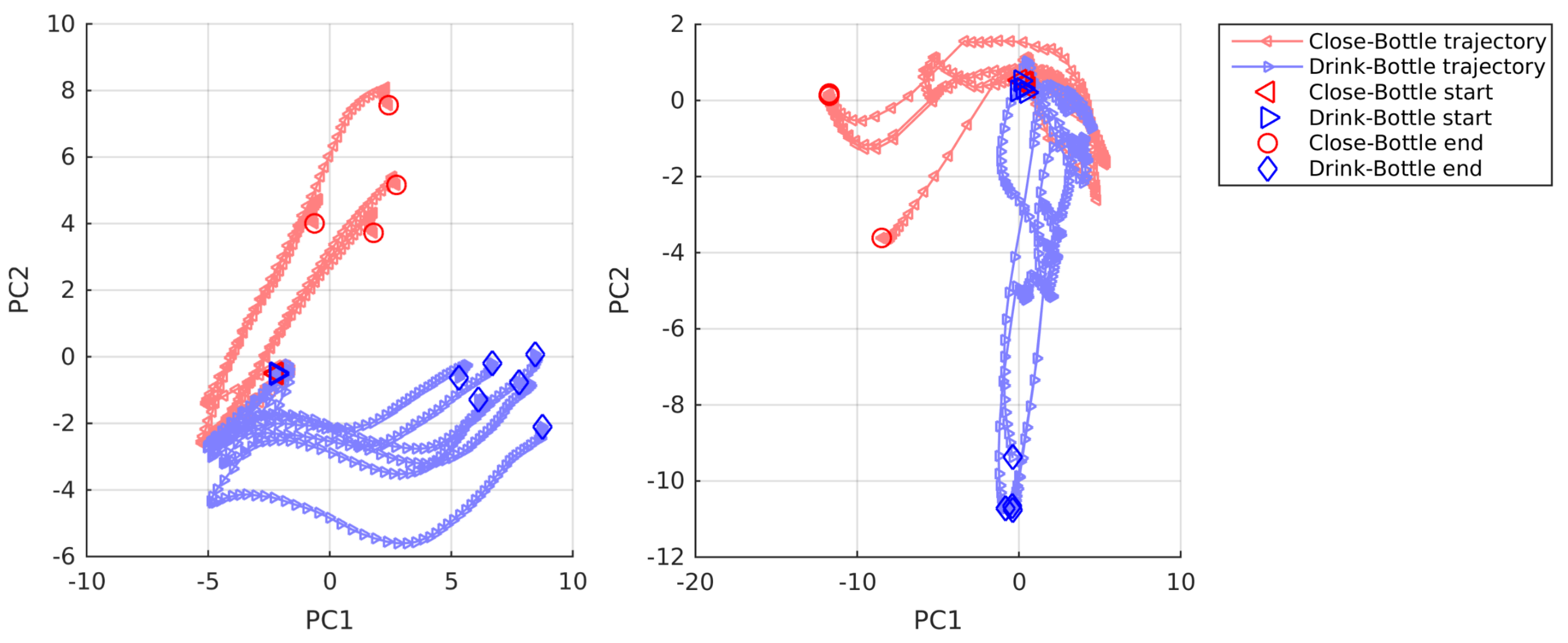}
	\centering
	\\ \qquad  (A) \qquad \quad\qquad\qquad\qquad\qquad\qquad\qquad(B) \qquad\qquad\qquad\qquad\qquad\qquad
	\caption{{PCA trajectories obtained from correctly categorized cases of Close/Drink-Bottle.}
		(A) Trajectories of the MSTRNN. (B) Trajectories of the LRCN.}
	\label{fig7}
\end{figure*}

The PCA trajectories of the LRCN (Fig. 7 (B)) also seem to share paths in the early time steps and differentiate at certain points, like the trajectories by the MSTRNN. However, their trajectories seem to make fluctuated movements, sometimes even going back and forth. This phenomenon may have occurred because the LSTM could not learn to develop temporal hierarchy because it does not have any temporal constraints imposed on its learning process. This may be the reason why the categorization performance of the LRCN is lower than the MSTRNN.

\subsection{Categorization of Compositionality Level 2 Action Dataset}

In the third experiment, the categorization performances of MSTRNN, MSTNN, and LRCN were compared by introducing a more challenging task with using the compositionality level 2 action dataset (CL2AD). Each action pattern in CL2AD can be expressed by an object, an action, and an action modifier.

\textbf{Dataset}: CL2AD consists of 840 videos that can be tagged by labels of objects, actions, and action modifiers. The categorization of action modifiers often requires the action recognition models to extract longer temporal correlation  than the recognition of ADOs or actions. For example, for a video of Book-Touch-2 Times category, because the model can acquire sufficient information only after perceiving a book twice touched, the model needs to extract a long-ranged temporal correlation between two similar events, the first touch and the second touch in order to identify the proper modifier as “twice”. In Fig. 8, we show the sampled images from CL2AD dataset. With a static image, it is hard to infer the action. But it is even harder to infer the action modifier.

Each video in the CL2AD are describable by compositions of four objects, four actions, and six action modifiers. Assumed categories in the dataset in terms of the object-action-modifier triplets are shown in Fig. 9. The total number of triplets is 42. The video recordings were taken from 10 different subjects. A subject shot two videos for each object, action, and modifier triplet. During dataset recording, subjects generated each action naturally. In a video, a non-ADO appears alongside with an ADO. A non-ADO was randomly picked from ADO categories excluding the true ADO of the video.  Objects were located in various positions in the task space. However, the camera view angle was fixed (the problem of view invariance is beyond the scope of the current study). The dataset is open for a public use (available at \url{http://neurorobot.kaist.ac.kr/programs.html}).

			\begin{figure}[!h]
				
				\includegraphics[scale=0.4]{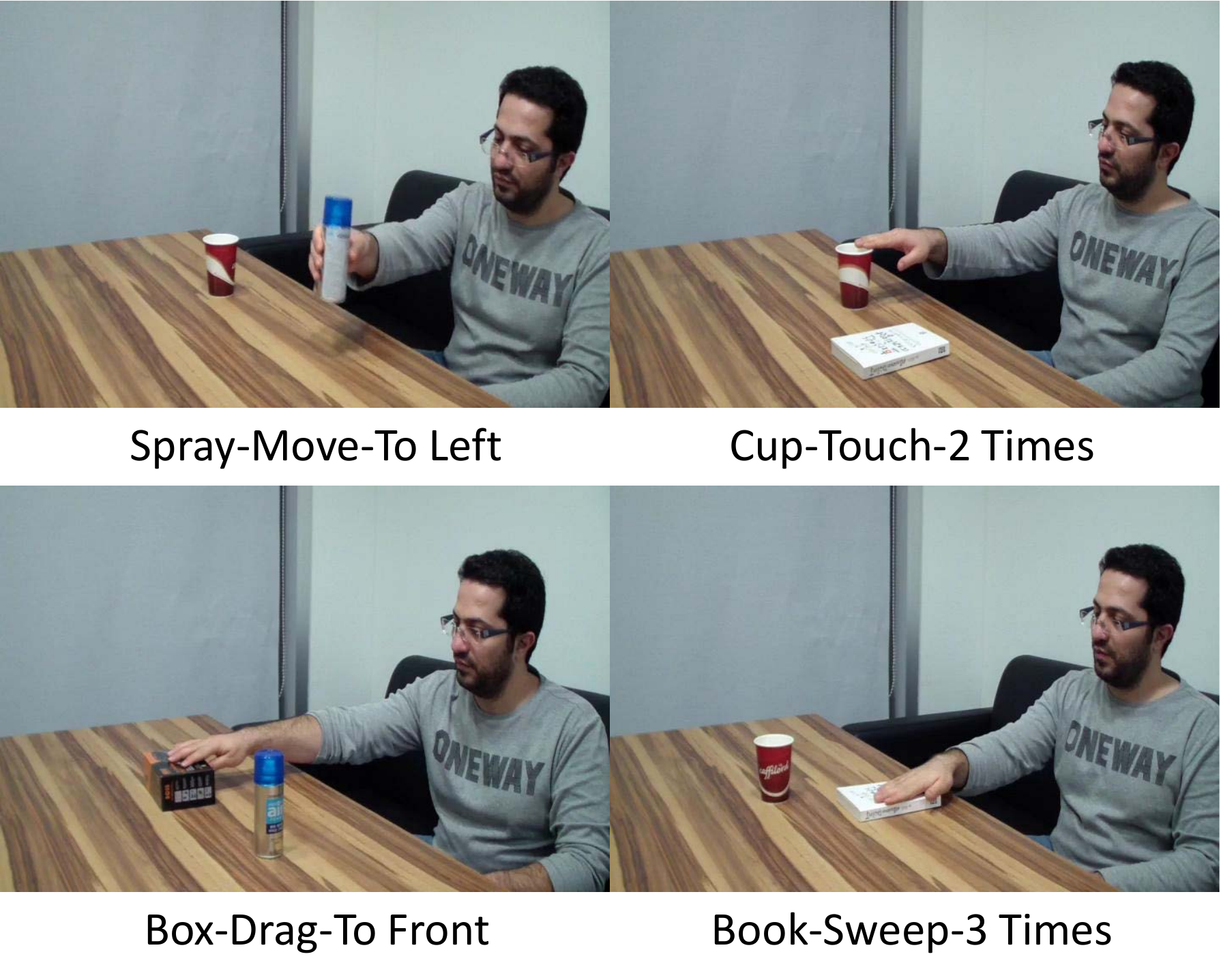}
				\centering
				\caption{{Sample frames from CL2AD. It is hard to infer the action category from a static image. But it is even harder to infer the action modifier from the image. Recognition of action modifiers requires a action recognition model to extract longer temporal correlation than the recognition of actions or ADOs.}}
				\label{fig8}
			\end{figure}	

			\begin{figure}[!h]
				
				\includegraphics[scale=0.4]{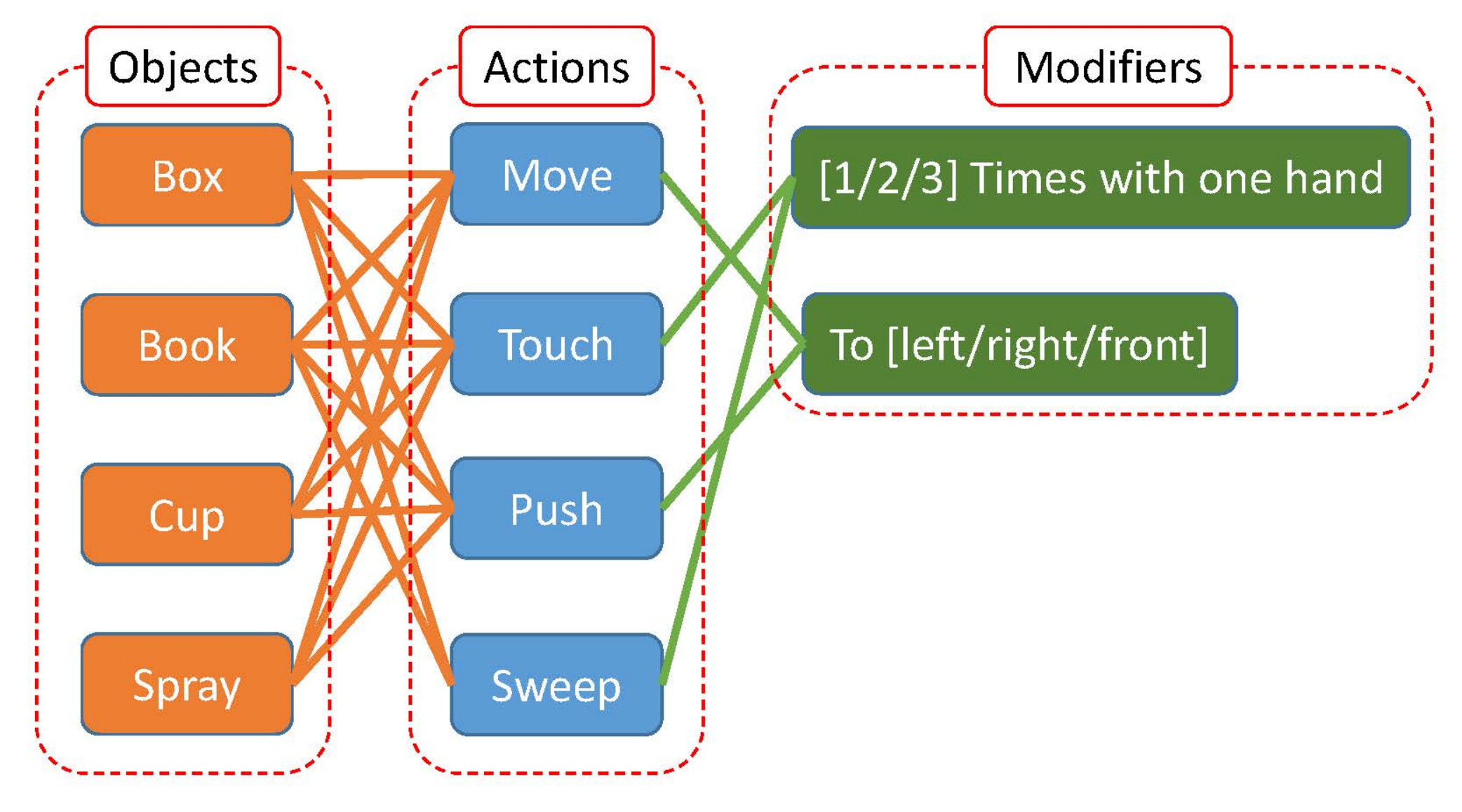}
				\centering
				\caption{{Composition of the CL2AD. Each object-action-modifier triplet or joint category is indicated by a connection. There are 42 joint categories.}}
				\label{fig8}
			\end{figure}

\textbf{Results}: The results obtained by testing the MSTRNN, MSTNN, and LRCN on the CL2AD are shown in Table V. The MSTRNN exhibited a higher modifier categorization performance than the MSTNN and LRCN, by 11.67\% and 12.74\% respectively. And for the ADO and action categorization accuracies, the MSTRNN performances were higher than the LRCN, but lower than the MSTNN by a small gap. 

The performance difference between the MSTRNN and MSTNN may have been caused by the enhanced categorical memories in the MSTRNN. As explained previously, the categorization of action modifiers requires more temporal information than the categorization of ADOs or actions. Therefore the MSTRNN, which has better categorical memories than the MSTNN, was able to achieve higher modifier categorization accuracy than the MSTNN. And the MSTNN may have shown higher accuracy in the ADO and action categorization than the MSTRNN because while its categorical memories are not effective for the modifier categorization, they were sufficient for the action categorization. Therefore the MSTNN may have used more of its resources in the ADO and action categorization than in the modifier categorization.

					\begin{table}[h]
						\centering
						\caption{
							{Recognition accuracies on the CL2AD in percentage}}
\begin{tabular}{crrr}
	\hline
	Category & \multicolumn{1}{c}{MSTRNN} & \multicolumn{1}{c}{MSTNN} & \multicolumn{1}{c}{LRCN} \\\hline
	ADO      & 75                         & 76.79                     & 68.1                     \\\hline
	Action   & 56.67                      & 61.67                     & 52.98                    \\\hline
	Modifier & 61.43                      & 49.76                     & 48.69 \\ \hline                  
\end{tabular}
						
						\label{table1}
					\end{table}

\section{Conclusions and Discussions}

We proposed a MSTRNN model for action recognition which extracts spatio-temporal features simultaneously by using multiscale spatio-temporal constraints imposed on the neural activity in different layers. The MSTRNN and MSTNN were compared by using the three actions concatenated patterns from Weizmann dataset in the first experiment. Results showed that due to the recurrent structure in the MSTRNN, the model outperformed the MSTNN. The comparative analysis on the neural activation sequences generated by both models suggest that the MSTRNN can develop more enhanced categorical memories by which compositional categorization of visually grounded data can be done more effectively than the MSTNN does. This is consistent with the biological evidences that a human brain utilizes its recurrent pathways to learn temporal sequences effectively \cite{goel2014timing}.

In the second experiment, the performance of the MSTRNN, MSTNN, and LRCN were compared by using compositionality level 1 action dataset, which is more challenging categorization task than the first one. The MSTRNN showed the highest categorization accuracies for both action-directed object (ADO) and action categories among the tested models. And the comparative analysis on the internal dynamics between the MSTRNN and LRCN suggested that the performance by MSTRNN is better than the one by LRCN. This is because the temporal hierarchy of the MSTRNN was successfully developed with temporal constraints adequately imposed on the model whereas the LRCN had no temporal constraints and could not form a hierarchy. 

The MSTRNN, MSTNN, and LRCN were again compared in the third experiment. In the experiment, compositionality level 2 action dataset which has triplets of ADOs, actions and modifiers as labels was used. In the results, the MSTRNN outperformed the other models in modifier categorization that requires the extraction of longer temporal correlation than the categorization of ADOs and actions. Both MSTRNN and MSTNN, which extract spatio-temporal features simultaneously, showed higher ADO categorization rate than the LRCN, which extracts spatial and temporal features only separately in different layers. This suggests that simultaneous extraction of spatial and temporal features are advantageous for learning-based action recognition models. This is consistent with the biological evidences that spatio-temporal receptive fields of a neuron in a mammalian cortex increases as the layer goes up \cite{s-t-field-mammal-increase1}, \cite{s-t-field-mammal-increase2}.

Although the MSTRNN showed better performance than the other models used for comparison, its recognition accuracy is far less than the human level. One of the main reasons for such degeneracy might be overfitting. In the future study, this overfitting problem may be alleviated with recently developed deep learning regularization techniques including the dropout technique for recurrent connections based on variational inference \cite{Dropout_RNN}. Recurrent batch normalization \cite{RecurrentBatchNormal} may also help to alleviate overfitting. Cooijmans et al. have shown that batch-normalization of LSTM improves its generalization capacity and encourages faster convergence in the learning phase.

Future study should also investigate on the possible advantage of adding a top-down prediction pathway and attention process to the MSTRNN structure. By adding a top-down pathway with attention, the recognition process should involve with dense interaction between the top-down proactive process of projecting possible image with attention and the bottom-up perceptual process of reflecting the top-down expectation by the perceptual reality. Such recognition process developed on such interactions between the both sides is considered to be more robust against possible perturbation in the input visual signals.

Finally, the future study should aim for further scaling. As mentioned previously, the current dataset used for the experiment 2 and 3 were built by the authors because there have been no adequate tagged video data with different levels of action compositionality in public datasets. It is, however, admitted that the size of the data is relatively small and this is why the model cannot be tested under more general conditions such as view angle free or size free conditions. Future study should explore how to enlarge data size efficiently for the purpose of scaling and generalization of the model system.

\section*{Acknowledgment}
This work was supported by the National Research Foundation of Korea (NRF) grant funded by the Korea government (MSIP) (No. 2014R1A2A2A01005491).

\ifCLASSOPTIONcaptionsoff
  \newpage
\fi

\begin{IEEEbiography}[{\includegraphics[width=1in,height=1.25in,clip,keepaspectratio]{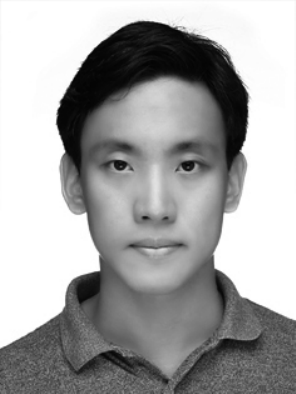}}]{Haanvid Lee} received the B.S. degree in electrical and electronic engineering from Yonsei University, Seoul, Republic of Korea, in 2015, and the M.S. degree in electrical engineering from Korea Advanced Institute of Science and Technology, Daejeon, Republic of  Korea, in 2017.

His current research interests include computer vision, recurrent neural networks, and reinforcement learning.

\end{IEEEbiography}

\begin{IEEEbiography}[{\includegraphics[width=1in,height=1.25in,clip,keepaspectratio]{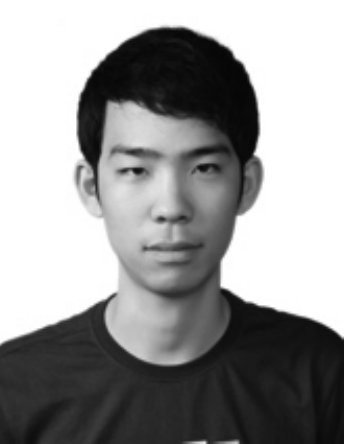}}]{Minju Jung} received the B.E. degree in electronic engineering from Hanyang University, Seoul, Republic of Korea, in 2013.
	
	His current research interests include recurrent neural networks and computer vision.
	
\end{IEEEbiography}

\begin{IEEEbiography}[{\includegraphics[width=1in,height=1.25in,clip,keepaspectratio]{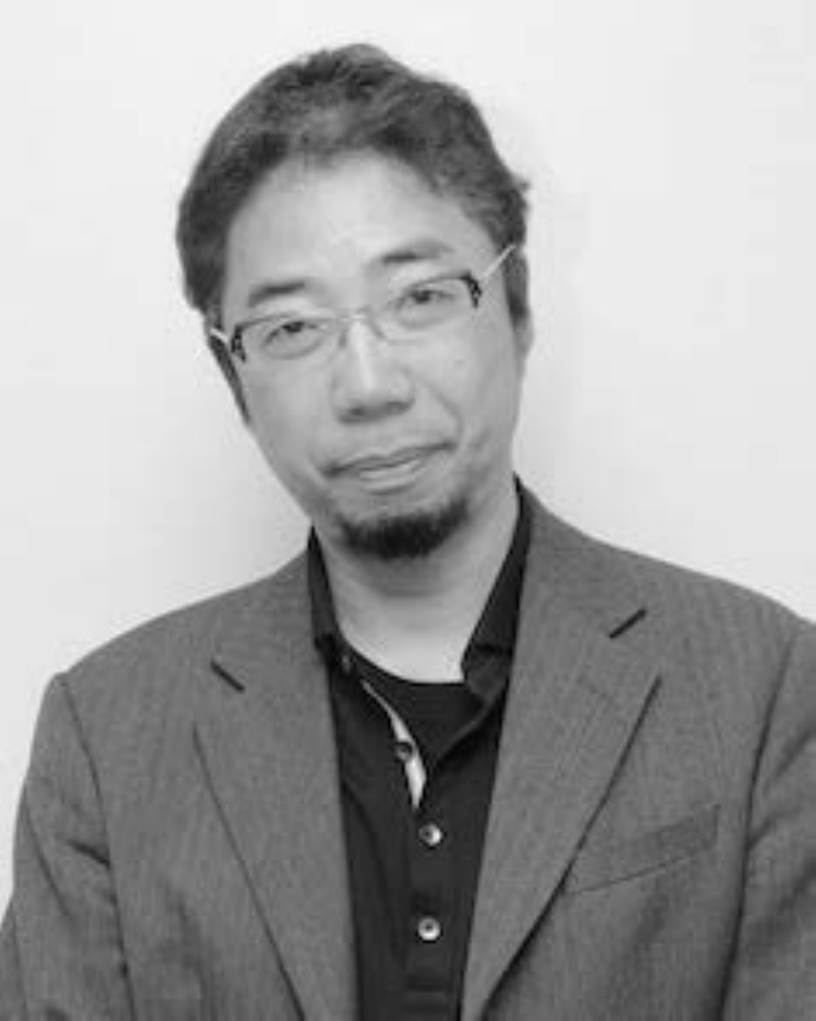}}]{Jun Tani} received the B.S. degree in mechanical engineering from Waseda University, Tokyo, Japan, the M.S. degrees in electrical engineering and mechanical engineering from the University of Michigan, Ann Arbor, MI, USA, and the D.Eng. degree from Sophia University, Tokyo.
	
He started his research career with the Sony Laboratory, Tokyo, Japan, in 1990. He had been a Team Leader of the Laboratory for Behavior and Dynamic Cognition, RIKEN Brain Science Institute, Saitama, Japan, for 12 years until 2012. He was a Visiting Associate Professor with the University of Tokyo, Tokyo, from 1997 to 2002. He is currently a Full Professor with the Electrical Engineering Department, Korea Advanced Institute of Science and Technology, Daejeon, Republic of  Korea. He is also currently a Adjunct Professor with the Okinawa Institute of Science and Technology, Okinawa, Japan. His current research interests include neuroscience, psychology, phenomenology, complex adaptive systems, and robotics.
	
\end{IEEEbiography}
\vfill\vfill\vfill\vfill\vfill\vfill\vfill







\begin{thebibliography}{1}
	
	
	\bibitem{CNN}
	Y.~LeCun, L.~Bottou, Y.~Bengio, and P.~Haffner, ``{Gradient-based learning
		applied to document recognition},'' \emph{Proceedings of the IEEE}, vol.~86,
	no.~11, pp. 2278--2323, 1998.
	
	\bibitem{CNN-ImageNet}
	J.~Deng, W.~Dong, R.~Socher, L.-J. Li, K.~Li, and L.~Fei-Fei, ``{ImageNet: A
		large-scale hierarchical image database},'' \emph{2009 IEEE Conference on
		Computer Vision and Pattern Recognition}, pp. 2--9, 2009.
	
	\bibitem{CNN-6.67}
	C.~Szegedy, W.~Liu, Y.~Jia, P.~Sermanet, S.~Reed, D.~Anguelov, D.~Erhan,
	V.~Vanhoucke, and A.~Rabinovich, ``{Going deeper with convolutions},'' in
	\emph{Proceedings of the IEEE Computer Society Conference on Computer Vision
		and Pattern Recognition}, vol. 07-12-June-2015, 2015, pp. 1--9.
	
	\bibitem{CNN-HumanLevel}
	O.~Russakovsky, J.~Deng, H.~Su, J.~Krause, S.~Satheesh, S.~Ma, Z.~Huang,
	A.~Karpathy, A.~Khosla, M.~Bernstein, A.~C. Berg, and L.~Fei-Fei, ``{ImageNet
		Large Scale Visual Recognition Challenge},'' \emph{International Journal of
		Computer Vision}, vol. 115, no.~3, pp. 211--252, 2015.
	
	\bibitem{3DCNN1}
	S.~Ji, W.~Xu, M.~Yang, and K.~Yu, ``3d convolutional neural networks for human
	action recognition,'' \emph{IEEE transactions on pattern analysis and machine
		intelligence}, vol.~35, no.~1, pp. 221--231, 2013.
	
	\bibitem{LRCN}
	J.~Donahue, L.~A. Hendricks, S.~Guadarrama, M.~Rohrbach, S.~Venugopalan,
	T.~Darrell, and K.~Saenko, ``{Long-term recurrent convolutional networks for
		visual recognition and description},'' in \emph{Proceedings of the IEEE
		Computer Society Conference on Computer Vision and Pattern Recognition}, vol.
	07-12-June-2015, 2015, pp. 2625--2634.
	
	\bibitem{ActionRecog1-TwoStream}
	K.~Simonyan and A.~Zisserman, ``Two-stream convolutional networks for action
	recognition in videos,'' in \emph{Advances in neural information processing
		systems}, 2014, pp. 568--576.
	
	\bibitem{LSTM}
	F.~A. Gers, N.~N. Schraudolph, and J.~Schmidhuber, ``{Learning Precise Timing
		with LSTM Recurrent Networks},'' \emph{Journal of Machine Learning Research},
	vol.~3, no.~1, pp. 115--143, 2002.
	
	\bibitem{s-t-field-mammal-increase1}
	D.~J. Felleman and D.~C. {Van Essen}, ``{Distributed hierarchical processing in
		the primate cerebral cortex},'' \emph{Cerebral Cortex}, vol.~1, no.~1, pp.
	1--47, 1991.
	
	\bibitem{s-t-field-mammal-increase2}
	U.~Hasson, E.~Yang, I.~Vallines, D.~J. Heeger, and N.~Rubin, ``{A hierarchy of
		temporal receptive windows in human cortex.}'' \emph{The Journal of
		neuroscience}, vol.~28, no.~10, pp. 2539--50, 2008.
	
	\bibitem{Downward-Causation}
	D.~T. Campbell, ``‘downward causation’in hierarchically organised
	biological systems,'' in \emph{Studies in the Philosophy of Biology}.\hskip
	1em plus 0.5em minus 0.4em\relax Macmillan Education UK, 1974, pp. 179--186.
	
	\bibitem{Downward-Causation-Gazzaniga}
	D.~S. Bassett and M.~S. Gazzaniga, ``Understanding complexity in the human
	brain,'' \emph{Trends in cognitive sciences}, vol.~15, no.~5, pp. 200--209,
	2011.
	
	\bibitem{MSTNN}
	M.~Jung, J.~Hwang, and J.~Tani, ``Self-organization of spatio-temporal
	hierarchy via learning of dynamic visual image patterns on action
	sequences,'' \emph{PloS one}, vol.~10, no.~7, p. e0131214, 2015.
	
	\bibitem{goel2014timing}
	A.~Goel and D.~V. Buonomano, ``Timing as an intrinsic property of neural
	networks: evidence from in vivo and in vitro experiments,'' \emph{Phil.
		Trans. R. Soc. B}, vol. 369, no. 1637, p. 20120460, 2014.
	
	\bibitem{Weizmann}
	L.~Gorelick, M.~Blank, E.~Shechtman, M.~Irani, and R.~Basri, ``{Actions as
		space-time shapes},'' \emph{IEEE Transactions on Pattern Analysis and Machine
		Intelligence}, vol.~29, no.~12, pp. 2247--2253, 2007.
	
	\bibitem{MTRNN}
	Y.~Yamashita and J.~Tani, ``Emergence of functional hierarchy in a multiple
	timescale neural network model: a humanoid robot experiment,'' \emph{PLoS
		Comput Biol}, vol.~4, no.~11, p. e1000220, 2008.
	
	\bibitem{LecunTanh}
	Y.~LeCun, L.~Bottou, and G.~Orr, ``Efficient backprop in neural networks:
	Tricks of the trade (orr, g. and m{\"u}ller, k., eds.),'' \emph{Lecture Notes
		in Computer Science}, vol. 1524.
	
	\bibitem{Elman-RNN}
	J.~L. Elman, ``Finding structure in time,'' \emph{Cognitive science}, vol.~14,
	no.~2, pp. 179--211, 1990.
	
	\bibitem{Jordan-RNN}
	M.~I. Jordan, ``Attractor dynamics and parallellism in a connectionist
	sequential machine,'' 1986.
	
	\bibitem{BPTT}
	D.~E. Rumelhart, J.~L. McClelland, P.~R. Group \emph{et~al.}, \emph{Parallel
		distributed processing}.\hskip 1em plus 0.5em minus 0.4em\relax IEEE, 1988,
	vol.~1.
	
	\bibitem{WeightDecay}
	A.~Krizhevsky, I.~Sutskever, and G.~E. Hinton, ``Imagenet classification with
	deep convolutional neural networks,'' in \emph{Advances in neural information
		processing systems}, 2012, pp. 1097--1105.
	
	\bibitem{Dropout}
	N.~Srivastava, G.~E. Hinton, A.~Krizhevsky, I.~Sutskever, and R.~Salakhutdinov,
	``Dropout: a simple way to prevent neural networks from overfitting.''
	\emph{Journal of Machine Learning Research}, vol.~15, no.~1, pp. 1929--1958,
	2014.
	
	\bibitem{3DCNN2}
	A.~Karpathy, G.~Toderici, S.~Shetty, T.~Leung, R.~Sukthankar, and L.~Fei-Fei,
	``Large-scale video classification with convolutional neural networks,'' in
	\emph{Proceedings of the IEEE conference on Computer Vision and Pattern
		Recognition}, 2014, pp. 1725--1732.
	
	\bibitem{small_kernel_best_simonyan2014very}
	K.~Simonyan and A.~Zisserman, ``Very deep convolutional networks for
	large-scale image recognition,'' \emph{arXiv preprint arXiv:1409.1556}, 2014.
	
	\bibitem{PCA}
	H.~Hotelling, ``Analysis of a complex of statistical variables into principal
	components.'' \emph{Journal of educational psychology}, vol.~24, no.~6, p.
	417, 1933.
	
	\bibitem{chinese_hand_writing_lstm}
	X.-Y. Zhang, F.~Yin, Y.-M. Zhang, C.-L. Liu, and Y.~Bengio, ``Drawing and
	recognizing chinese characters with recurrent neural network,'' \emph{arXiv
		preprint arXiv:1606.06539}, 2016.
	
	\bibitem{HandWritingBengio}
	J.~Chung, K.~Kastner, L.~Dinh, K.~Goel, A.~C. Courville, and Y.~Bengio, ``A
	recurrent latent variable model for sequential data,'' in \emph{Advances in
		neural information processing systems}, 2015, pp. 2980--2988.
	
	\bibitem{graves2013generating}
	A.~Graves, ``Generating sequences with recurrent neural networks,'' \emph{arXiv
		preprint arXiv:1308.0850}, 2013.
	
	\bibitem{Dropout_RNN}
	Y.~Gal and Z.~Ghahramani, ``A theoretically grounded application of dropout in
	recurrent neural networks,'' in \emph{Advances in Neural Information
		Processing Systems}, 2016, pp. 1019--1027.
	
	\bibitem{RecurrentBatchNormal}
	T.~Cooijmans, N.~Ballas, C.~Laurent, and A.~Courville, ``Recurrent batch
	normalization,'' \emph{arXiv preprint arXiv:1603.09025}, 2016.
	
\end{thebibliography}
\end{document}